\documentclass[runningheads]{llncs}

\usepackage{eccv}

\usepackage{eccvabbrv}

\usepackage{graphicx}
\usepackage{booktabs}
\usepackage{url}
\usepackage[accsupp]{axessibility}  % Improves PDF readability for those with disabilities.

\usepackage[pagebackref,breaklinks,colorlinks,citecolor=eccvblue]{hyperref}
\usepackage{orcidlink}

\usepackage{xcolor}
\usepackage{amsmath}
\usepackage{amssymb}
\usepackage{graphicx}
\usepackage{xspace}
\usepackage{multirow}
\usepackage{comment}
\usepackage{soul}
\usepackage{enumitem}

\definecolor{turquoise}{cmyk}{0.65,0,0.1,0.3}
\definecolor{purple}{rgb}{0.65,0,0.65}
\definecolor{dark_green}{rgb}{0, 0.5, 0}
\definecolor{orange}{rgb}{0.9, 0.6, 0.1}
\definecolor{red}{rgb}{0.8, 0.2, 0.2}
\definecolor{darkred}{rgb}{0.6, 0.1, 0.05}
\definecolor{blueish}{rgb}{0.0, 0.3, .6}
\definecolor{light_gray}{rgb}{0.7, 0.7, .7}
\definecolor{pink}{rgb}{1, 0, 1}
\definecolor{greyblue}{rgb}{0.25, 0.25, 1}

\begin{document}

% \title{Revisiting Token Dynamics in a Salience-Based Adaptive Masking for Enhanced Pre-training} 
\title{Salience-Based Adaptive Masking: Revisiting Token Dynamics for Enhanced Pre-training} 

\titlerunning{SBAM}

%\author{First Author\inst{1}\orcidlink{0000-1111-2222-3333} \and
%Second Author\inst{2,3}\orcidlink{1111-2222-3333-4444} \and
%Third Author\inst{3}\orcidlink{2222--3333-4444-5555}}
\author{Hyesong Choi\inst{1} \and Hyejin Park\inst{1} \and Kwang Moo Yi\inst{2} \and \\
Sungmin Cha\inst{3} \and Dongbo Min\inst{1}}

\authorrunning{H.~Choi \textit{et al}.}

%\institute{Princeton University, Princeton NJ 08544, USA \and
%Springer Heidelberg, Tiergartenstr.~17, 69121 Heidelberg, Germany
\institute{Ewha W. University \and University of British Columbia \and New York University}


\maketitle

\begin{abstract}
    %Mask-based pre-training has shown to be highly effective for training large models.
    %However, a common limitation shared by existing methods is their sensitivity to masking ratios.
    %In this work, we show that by focusing on token that are `salient', those that heavily impact reconstruction, one can create effective masks without any significant increase in computation cost.
    In this paper, we introduce Saliency-Based Adaptive Masking (SBAM), a novel and cost-effective approach that significantly enhances the pre-training performance of Masked Image Modeling (MIM) approaches by prioritizing token salience. 
    Our method provides robustness against variations in masking ratios, effectively mitigating the performance instability issues common in existing methods.
    This relaxes the sensitivity of MIM-based pre-training to masking ratios, which in turn allows us to propose an adaptive strategy for `tailored' masking ratios for each data sample, which no existing method can provide.
    Toward this goal, we propose an Adaptive Masking Ratio (AMR) strategy that dynamically adjusts the proportion of masking for the unique content of each image based on token salience. 
    We show that our method significantly improves over the state-of-the-art in mask-based pre-training on the ImageNet-1K dataset.
    %Code and model parameters are available at \url{https://sites.google.com/view/eccv24-sbam}.

  % Leading the vanguard of self-supervised learning (SSL), the triumphs of Masked Language Modeling (MLM) and its counterpart, Masked Image Modeling (MIM), stand as testaments to the paradigm's transformative potential.
  
\end{abstract}
\section{Introduction}
\label{sec:intro}

% \KY{Please see preamble.tex for how we typically collaborate over overleaf. You use the lowercase commands to color your edits so that others can see, and you use the uppercase commands for comments. Also remove colored edits as you go through and find okay}

Recent drastic improvements in various Computer Vision tasks rely heavily on Transformer architectures~\cite{dosovitskiy2020image}.
A critical component that enables these architectures is the necessity of large-scale data~\cite{deng2009imagenet}, which is not always readily available. 
Naturally, pre-training with pretext tasks has become a popular solution as a workaround, represented by Masked Image Modeling (MIM)~\cite{bao2021beit, he2022masked, xie2022simmim, dong2022bootstrapped}, inspired by how Masked Language Modeling (MLM)~\cite{devlin2018bert, liu2019roberta, clark2020electra, bao2020unilmv2} has reshaped the Natural Language Processing landscape.
These strategies involve masking a subset of the input data and predicting those that have been hidden, thus forcing the deep network to infer underlying concepts.
%A critical question then is how one designs these masks, as the quality of the pre-trained outcome is directly dependent on the masking strategy.

% In recent years, the landscape of self-supervised learning (SSL) has undergone a profound transformation, propelled by the integration of transformer architectures, a methodology first pioneered within the realm of natural language processing. This paradigm shift has not only revolutionized the field but has also paved the way for its application in computer vision, notably through the introduction of the Vision Transformer (ViT). Meanwhile, the inherent data-hungry nature of transformers has led to a growing consensus on the necessity of pre-training these models on pretext tasks, derived purely from the abundance of unlabelled data. A key strategy in this context is the application of masked token modeling, akin to Masked Language Modeling (MLM) employed in NLP. This strategy, adopted as Masked Image Modeling (MIM) for visual data, involves masking a subset of input tokens and training the model to predict these corrupted segments, thereby facilitating a deep contextual comprehension of the unlabelled data.

While MIM has significantly advanced self-supervised learning in vision, its conventional approach to randomly selecting tokens for masking falls short of harnessing the full potential of visual data~\cite{kakogeorgiou2022hide}. Unlike in text, where randomness might obscure key semantic units, the visual domain's complexity and token redundancy demand a more strategic masking protocol to ensure model comprehension. This necessity prompts us to explore a refined masking methodology, targeting a selection process that achieves the goal of image understanding for pre-trained models, thereby bridging the gap in modality-specific pre-training strategies.

% %Our endeavor seeks to meticulously evaluate and innovate upon the token masking process, ensuring that each decision to mask is as informed and impactful as possible.

%Various techniques have thus been proposed for more effective masking---in other terms, better pretext tasks for pre-training.

Various techniques~\cite{kakogeorgiou2022hide, wu2022object, liu2023good} have thus been proposed for more effective masking.
While these methods all strive toward improved pre-training, an oversight shared amongst all methods is that they do \emph{not} regard the contribution of each token within the overall composition of the image. It is crucial to scrutinize the interconnections between tokens to ensure that the token masking includes the tokens that play a pivotal role within the image. Moreover, prior masking methodologies~~\cite{kakogeorgiou2022hide, wu2022object, liu2023good} fail to address the critical consideration of the masking ratio, a factor that ought to dynamically adjust in response to the size and quantity of pivotal objects embedded within the image. This is primarily due to the difficulty reported in various works~\cite{he2022masked, xie2022simmim, yi2022masked} that minor changes in the optimized masking ratio can lead to performance instability, rendering such considerations difficult to implement. Thus, these methods must rely completely on chance and carefully tuned masking ratio to guarantee effective masking. Besides, contemporary masking strategies~\cite{kakogeorgiou2022hide, wu2022object, liu2023good} often grapple with high complexity, burdened by intricate distillation frameworks~\cite{kakogeorgiou2022hide}, auxiliary detection processes~\cite{wu2022object}, and duplicated forward processes~\cite{liu2023good}.%이 내용을 sec.9?에서 더 확인해라 추가하기

%how each masked patch (token) affects generation.
%This is critical, as in practice, each method requires a highly specific masking ratio for each dataset, to ensure that the masking process masks out essential parts of the image making the pre-training task non-trivial.
%For example, when learning to reconstruct a horse image, it is essential that masks are on the horse, otherwise the task becomes trivial.
%As existing methods do not consider how each token affects the final generation outcome, one has to rely completely on chance and carefully tuned masking ratio to guarantee effective masking.

In this work, we introduce a simple yet novel approach that focuses on token dynamics, termed \textbf{S}alience-\textbf{B}ased \textbf{A}daptive \textbf{M}asking (\textbf{SBAM}), which aims to strategically select masking locations by discerning perceptual prominence within the visual data. Specifically, the proposed method leverages the directional emphasis from attention mechanisms to identify the image tokens pivotal to the visual context.
Significantly, our methodology stands apart from the conventional attention-based approach~\cite{liu2023good} by leveraging the token's outgoing weight to calculate its `token salience' within the image and prioritizing those that have high salience to be masked. We also infuse a degree of randomness into token salience to enrich the diversity of mask generation. Further details of SBAM are elaborated in Sec.~\ref{sec:method_SBAM}. Thus, without any significant additional cost, we can consider the token's prominence within the image.

\begin{figure}[t]
    \centering
    \includegraphics[width=0.9\linewidth]{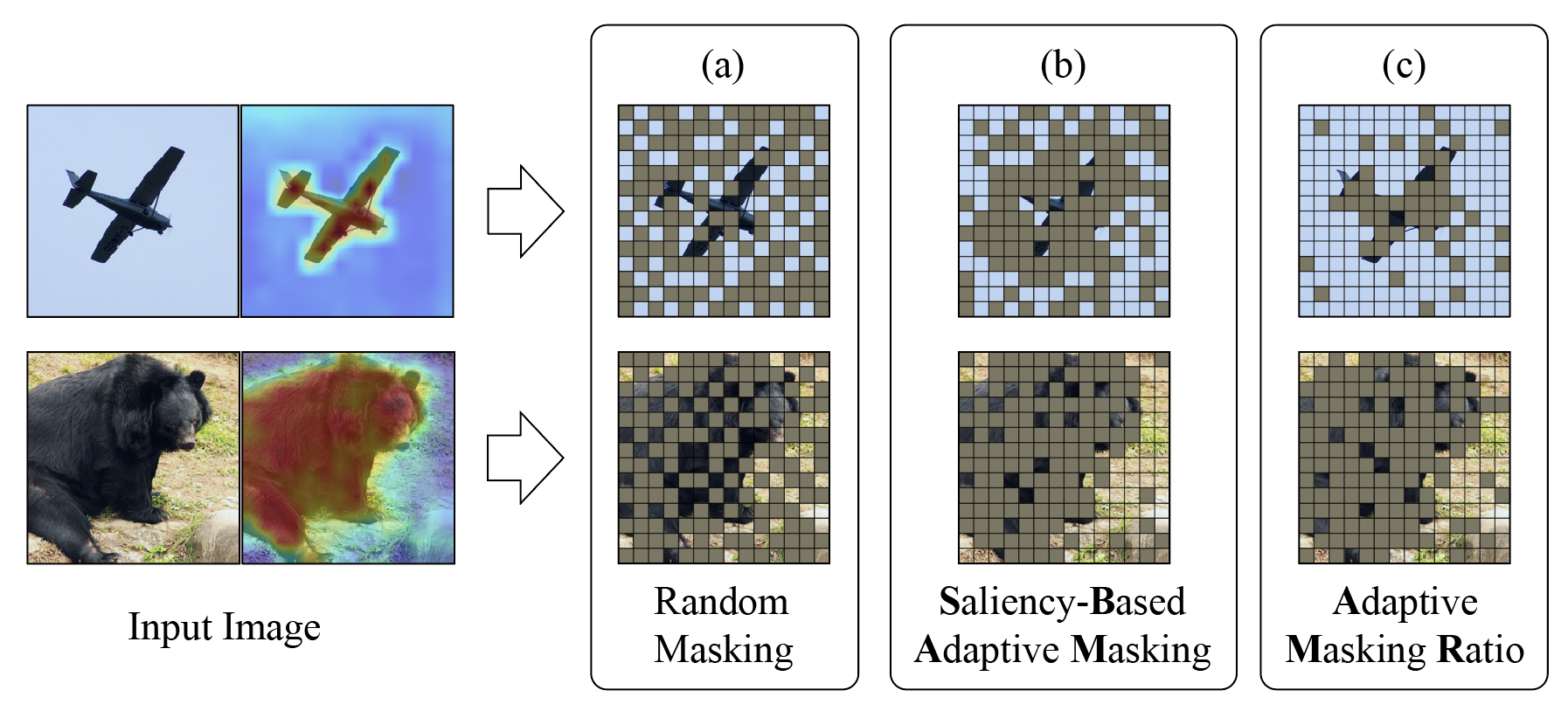}
    \caption{Overview of SBAM. Whereas (a) Random Masking must rely completely on chance and carefully tuned masking ratio to guarantee effective masking, (b) the proposed SBAM strategically masks tokens based on the token salience. The robustness of SBAM paved the way for the introduction of (c) AMR, which implements a tailored masking ratio for each sample in the dataset.}
    \label{fig:teaser}
    \vspace{-17pt}
\end{figure}

Crucially, the brilliance of the SBAM approach lies in its robustness to the varying masking ratios—a notable vulnerability in established baselines~\cite{he2022masked, xie2022simmim, yi2022masked}, which struggles from performance instability with even minor variations in optimized masking ratio. The robustness of SBAM is attributable to the proposed analytical precision in token dynamics, selectively masking tokens pivotal to the image's entirety. As a result, SBAM exhibits a diminished likelihood of masking redundant or non-essential segments, offering a stable alternative to the random masking or preceding masking approaches. Along with improving robustness to mask ratio variations, SBAM significantly enhances pre-training efficiency. A comprehensive evaluation of this is provided in Sec.~\ref{sec:SBAM_eval}.

% %By concentrating on the significance of tokens, as gauged by their outgoing weights, this approach presents a robust and flexible alternative to conventional methods, adeptly navigating the challenges posed by masking ratio variability.
% %This robustness facilitates a paradigm shift in MIM's approach to self-supervised learning, significantly enhancing its adaptability and performance across varying conditions.

Establishing robustness against variations in masking ratios has empowered us to expand the discourse on image masking into a pioneering aspect, introducing an innovative paradigm: an \textbf{A}daptive \textbf{M}asking \textbf{R}atio (\textbf{AMR}). We find having masking ratios that \emph{adapt} throughout training to be highly effective, as it allows the masking process to be \emph{tailored} to each sample in the dataset. For instance, each image may benefit from different masking ratios, as one image might have a close-up of a bear which would enjoy high masking ratios, whereas one might have an airplane in the sky that would require a lower masking ratio; see (c) of Fig.~\ref{fig:teaser}. The proposed token salience, reapplied in this context, serves to finely determine the dynamic masking ratio by quantifying the proportion of tokens exhibiting high salience. More details can be found in Sec.~\ref{sec:AMR_method}. The proposed AMR implements an adaptive strategy that respects the distinctiveness of visual data and thus achieves significant performance gains when applied to various baselines (refer to Sec.~\ref{sec:AMR_eval} and \ref{sec:AMR_baseline}. We note that this type of `tailoring' to each image for pre-training is impossible with any existing method.

%This leap forward acknowledges the inherently diverse visual narratives presented by individual images and adjusts the masking ratio to the varying objects and class sizes.
% %significant performance improvements, marking a significant step forward in the field of self-supervised learning.

In sum, this paper offers a pioneering masking strategy that enhances the robustness and effectiveness of pre-trained models, setting the stage for a paradigmatic shift in the field of self-supervised learning. More importantly, the proposed method can be universally applied across any MIM framework that exploits token masking. We evaluate our SBAM and AMR on data-hungry models like ViT-L/ViT-B~\cite{dosovitskiy2020image} on ImageNet-1K~\cite{deng2009imagenet} datasets, achieving significant performance improvements in both fine-tuning and linear probing accuracy.
% %Our contributions, through the introduction of a simplified yet profoundly effective masking strategy, not only bolster the resilience and adaptability of models like MIM and MAE but also pave the way for a transformative shift in self-supervised learning paradigms, with the potential to redefine the landscape of the field. and setting a new standard for adaptability and performance efficiency.

To summarize, our contributions are:
\vspace{-10pt}
\begin{itemize}
\setlength\itemsep{0em}
  \item we present Saliency-Based Adaptive Masking (SBAM), a novel effective method for MIM pre-training that focuses on token salience;
  \item being saliency-based, without a significant increase in computation, we allow effective masking that is robust to masking ratios;
  \item empowered by the robustness to masking ratio, we propose an Adaptive Masking Ratio (AMR) that allows \emph{tailored} masking for each sample;
  \item we evaluate our method on ImageNet-1K datasets, achieving notable enhancements in both fine-tuning and linear probing accuracy.
\end{itemize}

% \end{enumerate}
\section{Preliminaries}
Within the Masked Image Modeling (MIM) domain, the crux lies in the strategic corruption and subsequent reconstruction of image segments. This process hinges on two core operations: \textit{Random Masking} of image tokens and \textit{Reconstruction} of corrupted tokens to guide the learning process.

For a given image sequence $X \in \mathbb{R}^{N \times L \times D}$, where $N$, $L$ and $D$ denote batch size, number of tokens per image and dimensionality of each token respectively, and specified masking proportion $\gamma$, the general random masking process conducted by $\Phi_{\text{mask}}$ is defined as:
% \begin{equation}
% X_{\text{masked}}, M = \Phi_{mask}(X, \gamma),
% \end{equation}
\begin{equation}
X_{\text{masked}}, M = \Phi_{\text{mask}}(X, \gamma),
\end{equation}

%\begin{equation}
%X^{masked} = \Phi(X;\gamma) \oplus \Omega, \quad\quad\quad M = 1-\mathsf{B}(L,\gamma),
%\end{equation}

\noindent where $X_{\text{masked}}$ represents the visible tokens as a result of the post-application of the random mask $M$, with $M \in \{0,1\}^{N \times L}$ indicating the presence ($1$) or absence ($0$) of masking for each element.

% The reconstruction error $\mathcal{L}_{mim}$, fundamental to learning in MIM, is computed as the mean squared error (MSE) between the predicted representation of masked tokens $\hat{X}$ and their original counterparts $X$, normalized per the variance:
The key component for learning in MIM is the reconstruction error $\mathcal{L}_{\text{MIM}}$, which is computed by the mean squared error (MSE) between the predicted representation of masked tokens, denoted as $\hat{X}$, and their normalized original counterparts, represented by $X$:
% \begin{equation}
% X_{norm} = \frac{X-\mu(X)}{\sigma(X)+\epsilon}, \quad \mathcal{L}_{mim} = \frac{1}{\sum M} \sum\limits^{N}_{i=1} \sum\limits^{L}_{j=1} M_{ij} \cdot \Vert \hat{X}_{ij} - X_{norm, ij} {\Vert}^2_2.
% \end{equation}
\begin{equation}
\bar{X} = \frac{X-\mu(X)}{\sigma(X)+\epsilon}, \quad \mathcal{L}_{\text{MIM}} = \frac{1}{\sum_{i,j} M_{i,j}} \sum\limits^{N}_{i=1} \sum\limits^{L}_{j=1} M_{i, j} \cdot \Vert \hat{X}_{i, j} - \bar{X}_{i, j} {\Vert}^2_2.
\end{equation}

\noindent Here, $\mu(\cdot)$ and $\sigma(\cdot)$ represent the mean and standard deviation of $X$ respectively. $\Vert \cdot \Vert^2_2$ denotes the squared $L_2$ norm and $\epsilon$ is a small value for numerical stability.
This formulation presents the core of MIM's training objective, focusing explicitly on the reconstruction of the masked portions of the input, encouraging the model to infer corrupted information from the unmasked context.

\section{Salience-Based Adaptive Masking (SBAM)}~\label{sec:method_SBAM}
Masked Image Modeling (MIM) has greatly pushed forward self-supervised learning in the visual domain, yet its standard practice of randomly masking tokens fails to fully capture the richness of visual information. To overcome this limitation, we present Salience-Based Adaptive Masking (SBAM), a novel masking methodology that selects masking locations by revisiting token dynamics.

% of $X$ with its transpose, followed by a softmax \csm{function} to get the normalized affinities \csm{$\hat{\mathcal{A}}\in \mathbb{R}^{N \times L \times L \times N}$}. The softmax \csm{function $\rho(\cdot)$} is applied to \csm{$\mathcal{A}$} across the rows for each element $a_{i,j}$ in \csm{$\mathcal{A}$} as:

Given an input tensor $X \in \mathbb{R}^{N \times (L \times D)}$, the first step involves computing an affinity matrix $\mathcal{A}\in \mathbb{R}^{N \times L \times L}$ through a batch matrix-matrix product between $X$ and $X' \in \mathbb{R}^{N \times (D \times L)}$. Subsequently, we apply the softmax function $\rho(\cdot)$ to the affinity matrix, resulting in a normalized affinity matrix denoted as $\hat{\mathcal{A}} = \rho(\mathcal{A})\in \mathbb{R}^{N \times L \times L}$. The softmax function is implemented as follows:
% Namely, the softmax \csm{function} is applied as below:

\begin{equation}
% \csm{\mathcal{A}} = XX^T, \quad\quad\quad 
\rho({\mathcal{A}})_{n,i,j} =\frac{e^{a_{n,i,j}}}{\sum^L_{k=1}e^{a_{n,i, k}}}.
\end{equation}

Here, $a_{n,i,j}$ represents an element of $\mathcal{A}$ and $e$ is the base of the natural logarithm. Note that, for each $n$-th element with a shape of $\mathbb{R}^{L \times L}$, this function normalizes the rows of it, transforming them into probabilities that sum to 1.

% where $i$ ranges over all rows, $j$ and $k$ range over all columns, and $e$ is the base of the natural logarithm. This operation normalizes the rows of $A$, converting them into probability distributions that sum to 1.

% Given an input tensor \csm{$X \in \mathbb{R}^{N \times (L \times D)}$}, We first compute an affinity matrix \csm{$\mathcal{A}\in \mathbb{R}^{N \times N}$(?? please check this dimension is correct)} by performing a dot product of $X$ with its transpose.
% Then, we apply the softmax function $S(\cdot)$ to the affinity matrix to get the normalized affinities, such as \csm{$\hat{\mathcal{A}} = S(\mathcal{A})\in \mathbb{R}^{N \times N}$}.
% Namely, the softmax \csm{function} is applied across the rows of \csm{$\mathcal{A}$} as below:

% \begin{equation}
% \csm{\mathcal{A}} = XX^T, \quad\quad\quad \csm{S({\mathcal{A}})_{i,j}} =\frac{e^{a_{i,j}}}{\sum^L_{k=1}e^{a_{i, k}}},
% \end{equation}

% \csm{where $a_{i,j}$ denotes an element of $\mathcal{A}$ and $e$ is the base of the natural logarithm. This operation normalizes the rows of $A$, converting them into probability distributions that sum to 1.}

Crucially, our approach distinguishes itself from conventional attention-based methods by utilizing the sum of \textit{outgoing weight} of each token to determine the `token salience' $S \in \mathbb{R}^{N \times L}$ in the image, which is represented by the column-wise summed score of $\hat{\mathcal{A}}$:
\begin{equation}
S = \mathcal{N}(\Sigma^L_{j=1}\hat{\mathcal{A}}_{:,j,:}), \quad\quad\quad \mathcal{N}(x) = \frac{x-\text{min}(x)}{\text{max}(x)-\text{min}(x)}.
\end{equation}

In the affinity map, the row-wise score represents the incoming weight, reflecting the extent to which other tokens influence the corresponding token. Conversely, the column-wise score signifies the outgoing weight, indicating the contribution of the corresponding token to others. By summing these scores, it becomes feasible to quantify the token's overall impact on the image, thereby defining token salience.

The adaptive masking process of SBAM is formulated with the token salience $S$. Notably, exclusive reliance on $S$ for masking can precipitate a decline in performance (refer to Fig.~\ref{fig:ablations}). To mitigate this issue, we incorporated an element of randomness into the masking process to get the adjusted token salience, 
denoted as $\tilde{S} = S + N$, where $N \sim {U}([0,0.5)^{N\times L})$ represents a noise realization sampled from a multivariate uniform distribution $U$.
% denoted as $\tilde{S} = S + \eta$, where $\eta \in [0, 0.5)$ denotes random noise.

Then, the sampling of tokens for masking is guided by $\tilde{S}$. We sort $\tilde{S}$ in ascending order and select tokens corresponding to the top $K$ scores, where $K = \lceil L \cdot (1-\gamma) \rceil$. Consequently, an adaptive binary mask $M$ is constructed, where $M_i = 0$ for the $K$ selected tokens, and $M_i = 1$ for the remainder.

The proposed token salience offers a more intuitive and cost-efficient strategy for determining which tokens to mask. SBAM therefore revisits token dynamics in the image context, streamlining the conventional complex masking process and enabling strategic masking.

\begin{figure}[t]
    \centering
    \includegraphics[width=1\linewidth]{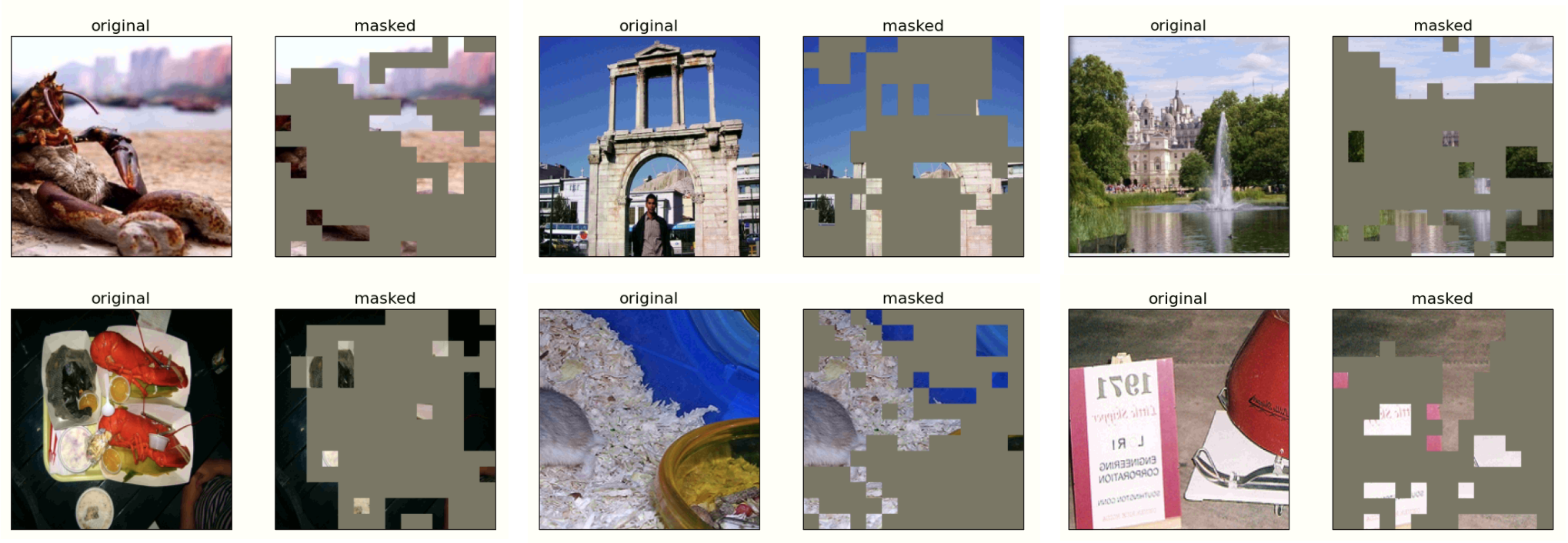}
    \caption{Qualitative example of SBAM. SBAM introduces `token salience' to prioritize and mask tokens with high significance. Hence, it is qualitatively confirmed that particularly important objects with high contribution within the image are selectively masked. Moreover, by integrating randomness with token salience, masks are probabilistically assigned to the background and less significant tokens, enriching the diversity of the token masking.}
    \label{fig:SBAM_qualitative}
    \vspace{-0.1cm}
\end{figure}

\section{Evaluation on SBAM}~\label{sec:SBAM_eval}
%robustness, fast pre-training
In this section, the evaluation critically examines the robustness of the proposed SBAM, especially against varying masking ratios, while also assessing its enhanced pre-training efficiency, evidenced by performance gains and faster convergence. Moreover, Fig.~\ref{fig:SBAM_qualitative} qualitatively shows that SBAM selectively masks only those tokens that contribute significantly to the image.

\subsection{Robustness to Masking Ratio Variability}
A crucial aspect of the evaluation of SBAM focuses on the robustness, particularly in the context of varying masking ratios. Prior established baselines, such as the widely used Masked Autoencoder (MAE)~\cite{he2022masked}, often exhibit significant performance fluctuations with even minimal adjustments to the masking ratio. This sensitivity undermines the practicality and generalizability of such methods, especially in diverse real-world scenarios where optimal masking ratios may not be consistent across datasets. In contrast, SBAM introduces a novel approach of selectively masking tokens based on their salience within an image. As a result, SBAM exhibits a diminished likelihood of masking redundant or trivial tokens, thereby maintaining a stable performance across a broad spectrum of masking ratios.

\begin{figure*}[t!]
\begin{center}
\includegraphics[width=0.99\linewidth]{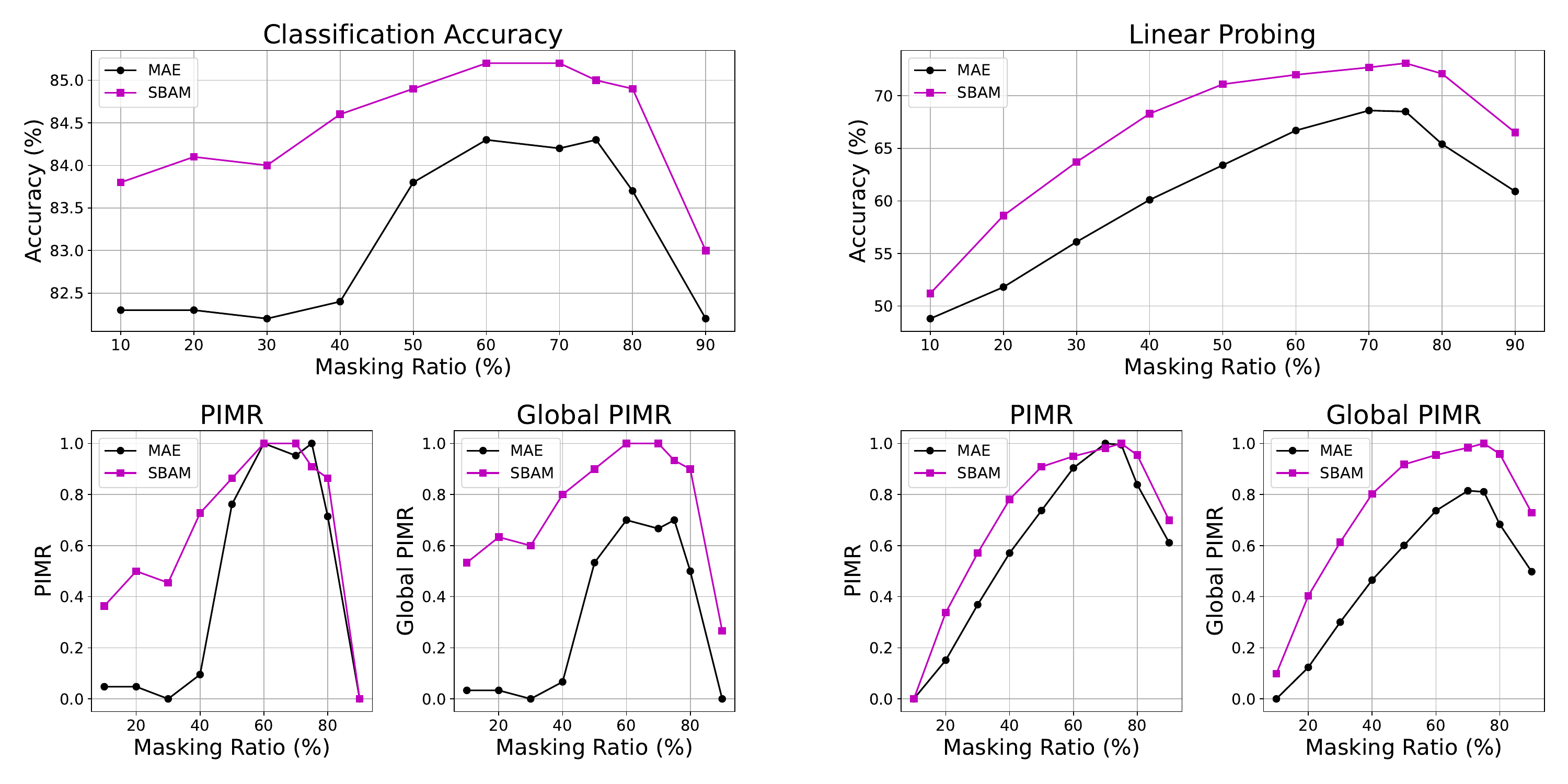}
\end{center}
\vspace{-15pt}
\caption{Evaluation of robustness across varied masking ratios. To evaluate the robustness of SBAM, we report the comparative analysis of image classification performance on ImageNet-1K dataset~\cite{deng2009imagenet} against the baseline method, MAE~\cite{he2022masked}, using ViT-L~\cite{dosovitskiy2020image} as a backbone. The upper graphs display the performance of the methods at different masking ratios, while the lower graphs illustrate the Performance Improvement over Masking Ratio (PIMR) and Global PIMR. These measures indicate the extent of each model's performance enhancement as the masking ratio increases from the lowest to higher ratios. SBAM significantly outperforms MAE in every measures, demonstrating its superior effectiveness in handling various masking ratios and enhanced pre-training performances.}
\label{fig:SBAM_robustness} 
\vspace{-0.1cm}
\end{figure*}

Fig.~\ref{fig:SBAM_robustness} showcases the performance stability of SBAM against strong baseline MAE, underscoring its superior resilience to changes in masking ratios. Our evaluation leverages the Performance Improvement over Masking Ratio (PIMR) metric, a normalized measure that quantifies how each model's performance at a given masking ratio stands against its performance at the lowest ratio, thereby reflecting the relative improvement. This metric is instrumental in revealing the impact of increased masked data on model training. 
The PIMR is defined as follows: 
\begin{equation}
\text{PIMR}(M) = \frac{P(M) - P(M_{min})}{P(M_{max}) - P(M_{min})},
\end{equation}
where $P(M)$ is the performance at masking ratio $M$. $P(M_{min})$ and $P(M_{max})$ denote the minimum and maximum observed performances, respectively. A PIMR value closer to 1 signifies a substantial improvement relative to the range of observed performances. The PIMR graph in classification accuracy demonstrate the robustness of the SBAM strategy, where it maintains a competitive edge over MAE across various masking ratios. Notably, SBAM shows a minimal performance drop at lower masking ratios compared to MAE, indicating its effectiveness even with sparse data presence. 

This advantage is particularly evident in the Global PIMR metric, where SBAM's performance remains consistently high.  For the Global PIMR calculation, we normalize performance relative to the most extensive range of performance observed among all models under comparison. This means that instead of comparing to the best and worst performances of a single model, Global PIMR considers the best and worst across both MAE and SBAM, which provides a universal performance context. Thus, the metric reflects a model's improvement not in isolation but rather in relation to its peers, which can be seen in the equation:
\begin{equation}
\text{Global PIMR}(M) = \frac{P(M) - P(M_{Gmin})}{P(M_{Gmax}) - P(M_{Gmin})},
\end{equation}
where $P(M_{Gmax})$ and $P(M_{Gmin})$ are the global minimum and maximum performance values observed across both MAE and SBAM models, respectively. This broader evaluation framework further underscores SBAM's superior resilience and ability to maintain high accuracy across varying degrees of masking, outperforming the MAE baseline even when less information is available for learning.

As shown in the graphs, SBAM consistently outperforms MAE across a spectrum of masking ratios, demonstrating its remarkable stability even at lower ratios. This is because in random masking, as the masking ratio decreases, the chance of including a crucial token in the mask lowers; conversely, SBAM consistently masks pivotal tokens, regardless of the masking ratio.

\begin{figure*}[t!]
\begin{center}
\includegraphics[width=0.99\linewidth]{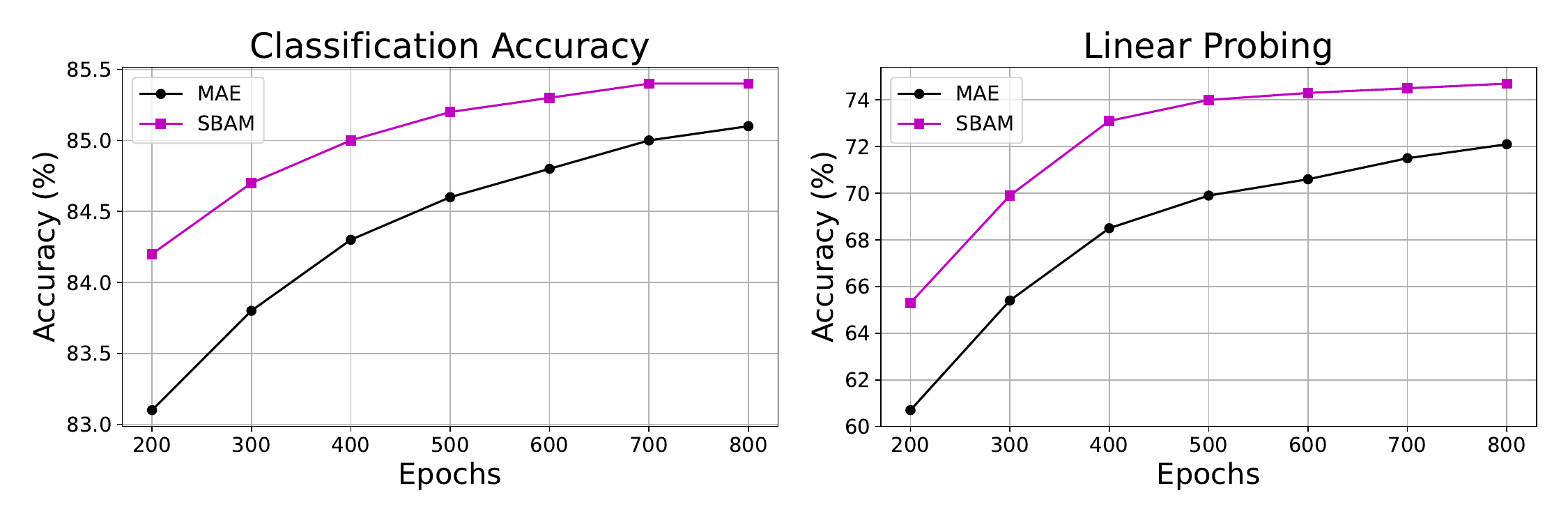}
\end{center}
\vspace{-15pt}
\caption{Performance evaluation of SBAM with respect to the pre-trained epochs. We report the comparison of image classification accuracy on ImageNet-1K~\cite{deng2009imagenet}, pre-trained on ViT-L~\cite{dosovitskiy2020image}. The left graph displays fine-tuning accuracy, whereas the right graph illustrates linear probing accuracy, both over a range of pre-trained epochs. The curves illustrate that SBAM surpasses MAE~\cite{he2022masked} in pre-training effectiveness in every trained epoch, and also validates its quicker attainment of converged performance levels.}
\label{fig:SBAM_performance}
\vspace{-0.4cm}
\end{figure*}

Furthermore, the classification accuracy graphs displayed above demonstrate that the application of SBAM results in a notable enhancement in performance compared to MAE for all ratios. This reveals that the proposed SBAM not only exhibits resilience to variations in the masking ratio but also significantly boosts pre-training efficacy, irrespective of the masking ratio. Strategically masking pivotal information enhances model performance and accelerates convergence by encouraging a comprehensive understanding of the visual context through a focus on essential tokens.

%To sum up, the observed robustness of SBAM is primarily attributed to its capability to mask out the most informative segments of an image, significantly reducing the chance of excluding critical visual information from training.

\subsection{Enhanced Pre-training Efficiency}
Beyond the robustness to masking ratio variations, SBAM's efficacy is further demonstrated through its enhanced pre-training capabilities. Traditional MIM strategies often require extensive computational resources and time for model convergence, primarily due to the indiscriminate masking of image tokens which can hinder the learning process by obfuscating essential visual cues. By strategically selecting pivotal tokens for masking, SBAM ensures that crucial tokens are leveraged during training, which leads to substantial improvement in pre-training performance and facilitates a more focused model convergence process.

Fig.~\ref{fig:SBAM_performance} showcases the comparative pre-training efficiency of SBAM against MAE~\cite{he2022masked}, underlining SBAM's reduced pre-training duration without a trade-off in accuracy. Initially, SBAM secures a distinct lead in classification accuracy and continues to demonstrate this advantage across the training epochs, as seen in the left graph. The right graph, representing linear probing accuracy, further confirms SBAM's higher initial performance, which stabilizes near peak levels well before 800 epochs. These findings highlight SBAM's ability to prioritize significant features during early training stages, resulting in a new alternative for both convergence speed and pre-training performance in the pre-training of masked image models.

% Figure~\ref{fig:figure2} illustrates the comparative pre-training speeds between SBAM and MAE, highlighting the substantial reductions in pre-training time achieved by SBAM without compromising model accuracy.
% %그래프의 경향성에 대해 언급하는 자세한 2-3문장 더 작성

% This efficiency stems from the method's ability to effectively prioritize learning on significant image features, thereby streamlining the pre-training phase and setting a new benchmark for speed and performance in masked image modeling.

\begin{figure}[t]
    \centering
    \includegraphics[width=1\linewidth]{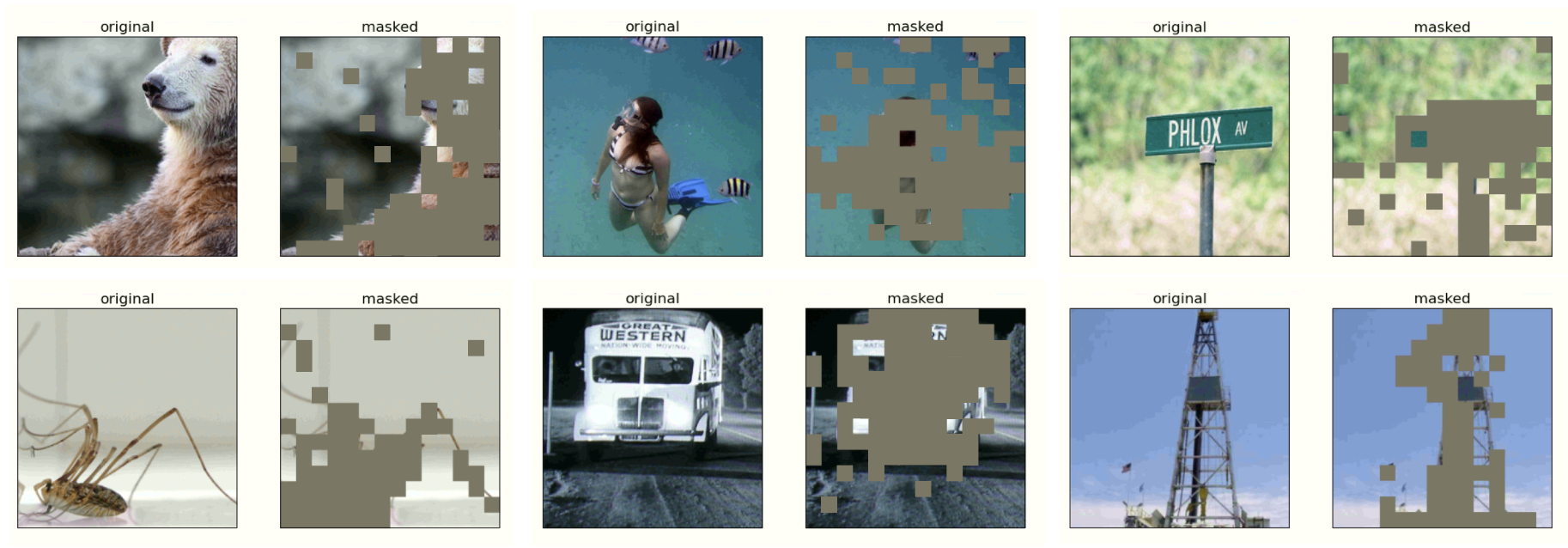}
    \caption{Qualitative example of AMR. Having masking ratios that adapt throughout training is highly effective, as it allows the masking process to be tailored to each sample in the dataset, accommodating the unique composition and object sizes within each image, as shown in the above qualitative samples.}
    \label{fig:AMR_qualitative}
    \vspace{-0.3cm}
\end{figure}

\section{Adaptive Masking Ratio (AMR)}~\label{sec:AMR_method}
Achieving stability against changes in masking ratios has enabled us to advance the discourse on image masking, introducing a novel perspective: an \textbf{A}daptive \textbf{M}asking \textbf{R}atio (\textbf{AMR}). This innovative approach acknowledges the inherently diverse visual narratives presented by individual images and adjusts the masking ratio to fit different object sizes and classes within them.

The proposed token salience $S = \mathcal{N}(\Sigma^L_{j=1}\hat{A}_{:,j,:})$ forms the basis for determining AMRs. The AMR $R_{dyna}$ is computed based on the distribution of salience scores across tokens, adjusted by a predefined variability parameter $\delta$ and the base mask ratio $r$:
\begin{equation}
R_{dyna} = r - \Delta r + 2 \Delta r \times \text{mean}(1_{S > \delta}).
\end{equation}

Here, $\Delta r$ denotes the range of allowable variation in the masking ratio, as $R_{dyna}$ can range from $r-\Delta r$ to $r+\Delta r$. $\delta$ is the salience threshold for distinguishing highly salient tokens, and 1 is the indicator function that identifies tokens exceeding $\delta$.

\begin{figure*}[t!]
\begin{center}
\includegraphics[width=0.99\linewidth]{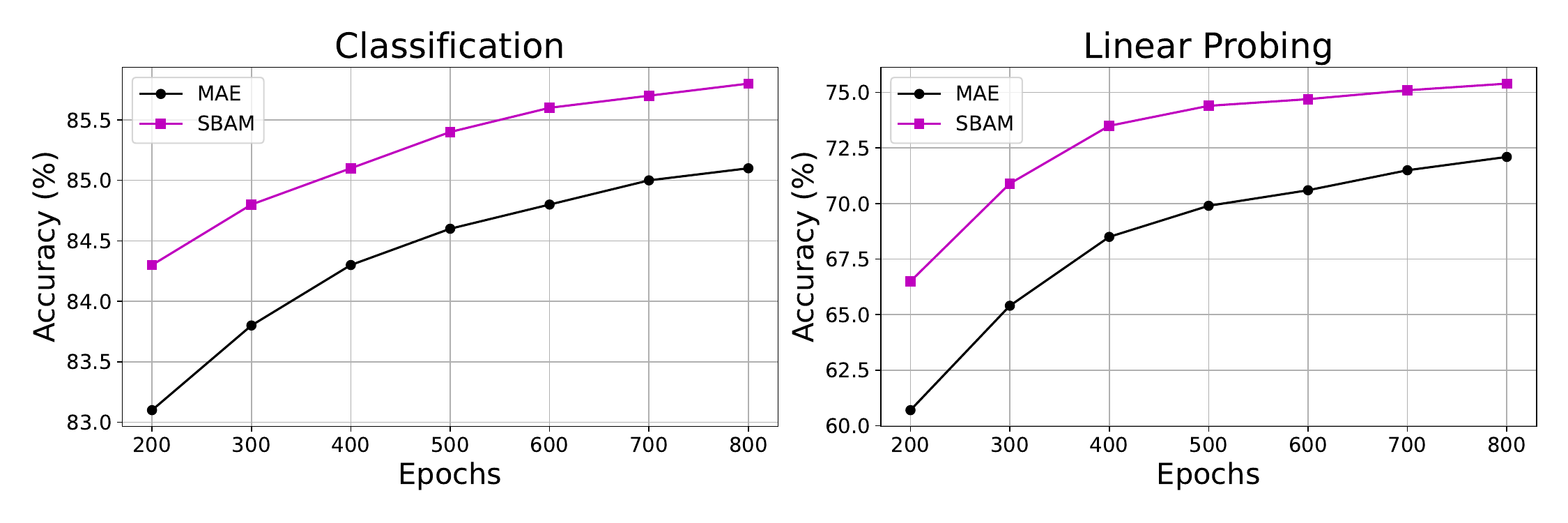}
\end{center}
\vspace{-10pt}
\caption{Performance comparison of AMR with respect to the pre-trained epochs. We report the comparison of image classification accuracy on ImageNet-1K~\cite{deng2009imagenet}, pre-trained on ViT-L~\cite{dosovitskiy2020image}. The left graph illustrates classification accuracy across epochs, while the right graph shows the accuracy obtained through linear probing. Both results indicate a significant improvement in pre-training performance when AMR is applied, which not only achieves higher accuracy earlier in the training process but also maintains a lead at convergence.} 
\label{fig:AMR_performance} 
\vspace{-5pt}
\end{figure*}

With $R_{dyna}$ established, we adjust the number of tokens to be masked accordingly. This dynamic adjustment of masking ratios ensures that the masking process is not uniformly applied but is instead sensitive to the visual information's inherent salience, promoting a more effective learning mechanism by focusing on the informative segments of the image. Fig.~\ref{fig:AMR_qualitative} presents a qualitative example of AMR, demonstrating the effectiveness of adaptive masking ratios that customize the masking process for each dataset sample, thereby accounting for the unique composition and object sizes within each image. As a result, the proposed AMR employs an adaptive approach, leading to enhanced performance across different models (See Fig.~\ref{fig:AMR_performance} and Tab.~\ref{tab:AMR_baselines}) and setting a new standard for tailored image masking.

\section{Experiments}
% enhanced converged performance of AMR on various baselines
% Ablation studies (hyperparameters)

\subsection{Implementation Details}
Our evaluation approach involves deploying the proposed SBAM and AMR against the baseline to assess performance enhancements. To ensure a fair comparison, we maintain consistency with the baseline method's hyperparameters and network architectures. Notably, to preserve the integrity of our experiments, we ensured uniform hardware utilization and experimental conditions across both our method and the reproduced baseline, all employing 8*A6000 GPUs. Given that this fixed GPU configuration diverges from those used in prior methods, discrepancies between our reproduced performance and the performance documented in existing papers may arise. For the fair experimental schedule, the reproduced 400-epoch performance of baseline methods and the proposed method were \textit{all equally measured in intermediate stages in the training towards 800 epochs}. For transparency, we released the model parameters for the main experiments along with the code on the anonymous project page. A more detailed implementation description can be found in the Supplementary material.

\subsection{Evaluation on AMR}~\label{sec:AMR_eval}
% Enhanced Pre-training Efficiency + enhanced converged performance (Fig. 4) 
The graphs in Fig.~\ref{fig:AMR_performance} demonstrate the efficacy of the AMR in comparison to the MAE model throughout the pre-training phase. Specifically, the left graph indicates that SBAM starts with a higher accuracy than MAE at the 200 epochs mark and continues to outperform MAE~\cite{he2022masked} at every subsequent checkpoint. By the 800 epochs mark, SBAM not only achieves a significant accuracy enhancement but also shows a more rapid improvement in the earlier epochs, suggesting that SBAM requires fewer epochs to achieve similar or better performance compared to MAE. In the context of linear probing, depicted in the right graph, the trend is similar. SBAM consistently achieves higher accuracy than MAE from the outset, and this performance gap is maintained as training progresses. The curves for SBAM and MAE become flat toward 800 epochs, indicating that 800 epochs or later is the point of convergence in performance. This observation underscores SBAM's substantial advantage over MAE, both in terms of converged performance and convergence speed. Overall, the performance trends captured in these graphs suggest that SBAM is more efficient during pre-training, reaching higher levels of accuracy faster than MAE.

\begin{table*}[t!]
\centering
\addtolength{\tabcolsep}{10pt}
\caption{Comprehensive performance results of applying SBAM to various baseline methods. We report the comparison of image classification fine-tuning accuracy on ImageNet-1K~\cite{deng2009imagenet} dataset. The consistent performance improvement of SBAM across various baseline methods demonstrates the efficacy of SBAM as a scalable methodology capable of enhancing a variety of MIM frameworks.}
\begin{tabular}{l|c|c}
\toprule
Method        & Baseline & Baseline+SBAM \\ \toprule \bottomrule
MAE (ViT-L)~\cite{he2022masked}  & 84.3	& 85.1 \\
MAE (ViT-B)           &82.9	& 83.6  \\
BootMAE (ViT-B)~\cite{dong2022bootstrapped}     & 84.1 &	84.8 \\
iBoT (ViT-B)~\cite{zhou2021ibot} & 71.5	& 74.4   \\
CMAE (ViT-B)~\cite{huang2023contrastive} & 83.8	& 84.5  \\ 
\bottomrule
\end{tabular}
% \vspace{-0.3cm}
\label{tab:AMR_baselines}
\end{table*}

\subsection{Evaluation on Various Baselines}~\label{sec:AMR_baseline}
% Comprehensive performance results of applying SBAM to various baseline methods (Tab. 1)
The integration of SBAM into various baseline methodologies demonstrates a notable enhancement in pre-training performance, as summarized in Tab.~\ref{tab:AMR_baselines}. All experiments report the classification accuracy on 400 epochs, except for iBoT~\cite{zhou2021ibot} which is pre-trained for 100 epochs. Applying SBAM to the large-scale variant of MAE~\cite{he2022masked} (MAE (ViT-L)) yields a noteworthy enhancement, elevating the baseline accuracy from 84.3\% to 85.1\%, marking a significant advancement. Integration of SBAM to the MAE using ViT-B also experienced significant performance gains, underscoring that the benefits of SBAM are not limited to specific model architectures and can provide substantial pre-training efficacy for a variety of models. The performance enhancement of SBAM on various models including BootMAE~\cite{dong2022bootstrapped}, iBoT, and CMAE~\cite{huang2023contrastive} highlights SBAM's generalizability and efficacy in augmenting various model structures with significant accuracy gains.

%The Bootstrapped version of MAE, BootMAE, which incorporates self-distillation, also shows improved performance with SBAM, indicating that SBAM's approach is complementary to methods leveraging internal consistency.

%Interestingly, iBoT (ViT-B, 100 epoch) shows a remarkable performance leap when enhanced with SBAM. The jump from 71.5\% to 74.4\% represents the model's newfound robustness and efficiency, highlighting SBAM's capability to bolster models with a focus on unsupervised learning.

%Lastly, CMAE (ViT-B), known for its contrastive masked autoencoder approach, benefits from SBAM with an accuracy increase to 84.5\%, underlining the versatility and applicability of SBAM across different pre-training strategies.

In conclusion, the consistent improvement across various baselines validates the efficacy of SBAM as a scalable enhancement tool. It not only boosts performance in standard settings but also bridges the gap in more challenging learning scenarios, marking it as a pivotal development in masked image modeling pre-training techniques.

\begin{table}[t!]
    \centering
    \begin{minipage}{.48\linewidth}
        \centering
        \caption{Comparative evaluation of SBAM against the state-of-the-art masking strategy AMT~\cite{liu2023good}.}
        \vspace{-5pt}
        \addtolength{\tabcolsep}{8pt}
        \begin{tabular}{l|c}
        \toprule
            Method                  & Acc (\%)  \\ \toprule \bottomrule
            AM~\cite{liu2023good}  & 82.5	 \\
            AMT~\cite{liu2023good}                     & 82.8  \\
            SBAM                    & 83.6 \\
            \bottomrule
        \end{tabular}
        \label{tab:table2}
    \end{minipage}%
    \hspace{0.15cm}
    \begin{minipage}{.48\linewidth}
        \centering
        \caption{Comparative evaluation of SBAM against the state-of-the-art masking strategy AttMask~\cite{kakogeorgiou2022hide}.}
        \vspace{-4pt}
        \addtolength{\tabcolsep}{8pt}
        \begin{tabular}{l|c}
        \toprule
            Method                  & Acc (\%)  \\ \toprule \bottomrule
            AttMask-High~\cite{kakogeorgiou2022hide}	&72.5	 \\
            AttMask-Hint~\cite{kakogeorgiou2022hide}	&72.8 \\
            SBAM	        &74.4 \\
            \bottomrule
        \end{tabular}
        \label{tab:table3}
    \end{minipage}
    \vspace{0.1cm}
\end{table}  

\begin{figure*}[t!]
\begin{center}
\includegraphics[width=0.99\linewidth]{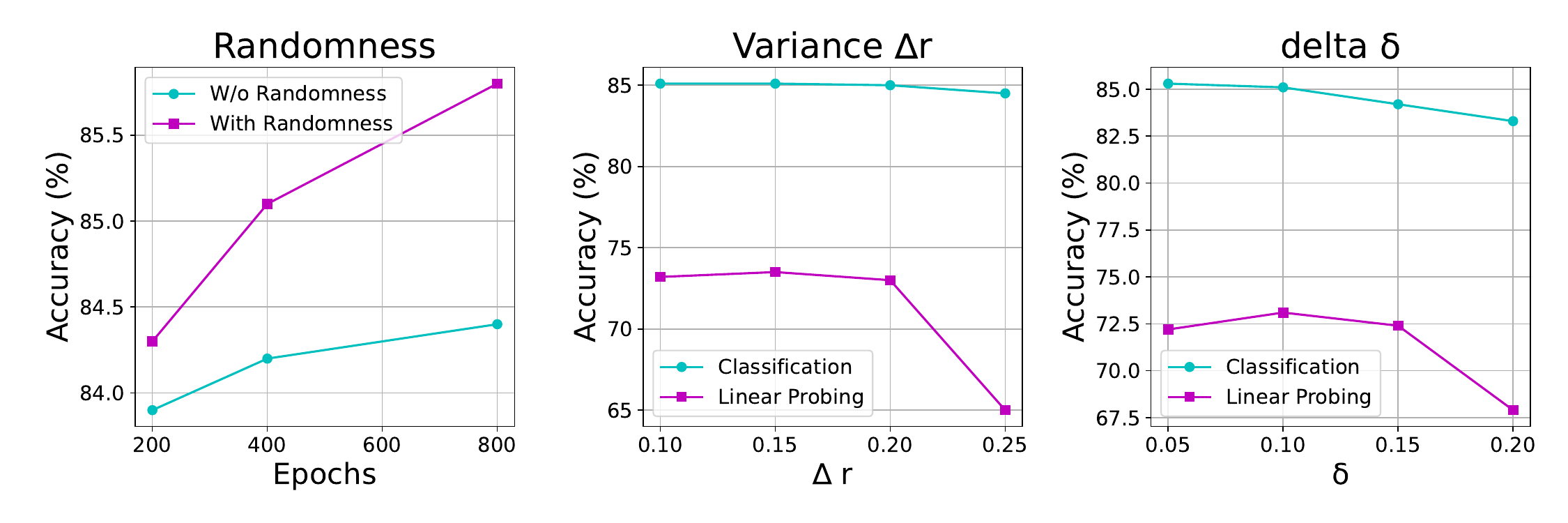}
\end{center}
\vspace{-10pt}
\caption{Comprehensive ablation studies of the impact of randomness, variance ($\Delta r$), and delta ($\delta$) of the proposed SBAM approach.}  
\label{fig:ablations} 
\vspace{0.2cm}
\end{figure*}

\vspace{0.2cm}
\subsection{Comparison with Masking Methods}
% (Tab2,3)
We conducted a comparative evaluation of our SBAM masking strategy against the established masking strategies of AMT and AttMask. This comparison was performed by applying SBAM to the baselines previously employed by AMT~\cite{liu2023good} and AttMask~\cite{kakogeorgiou2022hide} methods, specifically MAE~\cite{he2022masked} (ViT-B, 400 epochs) and iBoT~\cite{zhou2021ibot} (ViT-B, 100 epochs), to ascertain the enhancements introduced by our approach.
 
In Tab.~\ref{tab:table2}, the AM and AMT strategy~\cite{liu2023good}, which implement basic masking and selective masking based on semantic importance, achieve an image classification fine-tuning accuracy of 82.5 and 82.8, respectively. When applied to the same baseline model, SBAM outperforms both approaches by achieving an accuracy of 83.6. This implies that the efficacy of self-supervised pre-training can be maximized by defining token salience based on the outgoing weight of the token, as opposed to the incoming weight~\cite{liu2023good}.

Furthermore, we compare SBAM to two variants of AttMask~\cite{kakogeorgiou2022hide}, applied within the iBoT~\cite{zhou2021ibot} framework in Tab.~\ref{tab:table3}. Both AttMask-High and AttMask-Hint incorporate the selective masking strategy of distillation setup which leverages similarity to classification token. SBAM stands out with an accuracy of 74.4\%, substantially higher than AttMask-High's 72.5\% and AttMask-Hint's 72.8\%. This highlights the superiority of the SBAM method, which can effectively improve the pre-training efficiency without the need for the additional computational cost of using a complex framework.

\subsection{Ablation studies}
% (Fig. 5)
In Fig.~\ref{fig:ablations}, we provide comprehensive ablation studies of the impact of randomness, variance ($\Delta r$), and delta ($\delta$) of the proposed SBAM approach. We report the ImageNet-1K~\cite{deng2009imagenet} classification accuracy achieved by SBAM using the baseline approach of MAE~\cite{he2022masked}, trained for 400 epochs on ViT-L~\cite{dosovitskiy2020image} as the backbone. The first graph depicts the impact of the incorporated randomness in the SBAM on model performance over various epochs. The plot reveals that integrating randomness with token saliency markedly enhances pre-training accuracy as the number of pre-trained epochs increases. The second and third graphs show the fine-tuning accuracy and linear probing performance ablations for the hyperparameters of SBAM. While fine-tuning accuracy remained stable across various hyperparameters, linear probing accuracy demonstrated relative sensitivity. We chose $\Delta r = 0.15$ and $\delta = 0.1$ as optimal hyperparameters in both figures and were universally applicable across all baseline methodologies.

\section{Related Work}

\subsection{Masked Language Modeling}
Masked Language Modeling (MLM)~\cite{devlin2018bert, liu2019roberta, clark2020electra, bao2020unilmv2, zaken2021bitfit, ghazvininejad2019mask, song2020mpnet, song2019mass, raffel2020exploring, conneau2019cross} has become a keystone self-supervised learning paradigm in NLP, exemplified by groundbreaking models like BERT~\cite{devlin2018bert} and GPT~\cite{radford2018improving, radford2019language}. By predicting masked tokens from their context, MLM has propelled NLP forward, enabling models to scale and perform adeptly on diverse tasks~\cite{brown2020language}. However, the considerable training time and computational demands of these models have spurred innovations aimed at increasing pre-training efficiency. For example, ALBERT~\cite{lanalbert} reduced parameters through embedding matrix factorization and shared layer parameters, while EarlyBERT~\cite{chen2020earlybert} applied the principles of network pruning to optimize the training process. The curriculum masking approach of CCM~\cite{lee2022efficient} represents another advancement, strategically increasing the complexity of token masking to enhance learning. These efforts reflect a broader trend in the quest for efficiency, leading to models that maintain or exceed the capabilities of their predecessors with a fraction of the resource investment.

\subsection{Masked Image Modeling}

In computer vision, Masked Image Modeling (MIM))~\cite{chen2020generative, bao2021beit, zhou2021ibot, he2022masked, xie2022simmim, huang2022green, cao2022understand, dong2022bootstrapped, zhang2023hivit, liu2022mixmim, pan2022towards, zhang2022mask, peng2022beit, peng2022unified, hou2022milan, xue2023stare} has emerged as a transformative technique, drawing parallels to the success of Masked Language Modeling (MLM) in NLP. MIM's central tenet involves predicting occluded parts of images to foster a nuanced understanding of visual content sans explicit labels. Early efforts adapting MLM concepts for visual data, such as iGPT~\cite{chen2020generative}, paved the way for more sophisticated methods. BEiT~\cite{bao2021beit} utilized a pre-trained discrete variational autoencoder (dVAE) to produce target visual tokens. Further refinements in the technique have been observed in methods such as MAE~\cite{he2022masked} and SimMIM~\cite{xie2022simmim}, which focus on direct prediction from unmasked image areas, refining the process of visual understanding. Efficiency in pre-training has been a critical frontier, leading to innovations like GreenMIM~\cite{huang2022green} and HiViT~\cite{zhang2023hivit}, which optimize hierarchical Vision Transformers (ViTs) by processing only the visible patches, significantly reducing computational overhead.

Recent advances in model pre-training have honed in on the strategic use of masking to enhance learning efficiency, with a particular emphasis on selecting which image regions to mask. Initiating this trend, ADIOS~\cite{shi2022adversarial} leverages adversarial training to smartly select challenging segments for masking, setting a foundation for intelligent masking approaches. AttMask~\cite{kakogeorgiou2022hide} and SemMAE~\cite{li2022semmae} further this by utilizing self-attention and semantic information, respectively, to pinpoint and mask the most informative parts of an image, thereby prioritizing high-value areas over random masking. The Attention-Driven Masking and Throwing Strategy (AMT) strategy~\cite{liu2023good} refines this focus on semantics by employing self-attention to identify and eliminate redundant patches, achieving a delicate balance between precision and efficiency. In the realm of CLIP models, ACLIP~\cite{yang2023attentive} and Fast CLIP (FLIP)~\cite{li2023scaling} adopt attentive masking strategies to optimize training, with Fast CLIP demonstrating the effectiveness of masking substantial portions of images for accelerated learning.
MaskAlign~\cite{xue2023stare} introduces an innovative teacher-student framework that bypasses the need for masked region reconstruction, aligning visible features with semantically rich intact image features to concentrate on the most informative parts. Together, these approaches illustrate a shift towards more strategic, intelligent masking techniques, significantly boosting the pre-training process by leveraging both the quantity and quality of masked inputs.

\section{Conclusions}
%\vspace{-0.4cm}
The proposed Saliency-Based Adaptive Masking (SBAM) approach and the Adaptive Masking Ratio (AMR) significantly progress the field of MIM by introducing a method that adaptively masks image tokens with dynamic masking ratios based on their token salience. SBAM not only enhances pre-training efficiency and model performance on ImageNet-1K datasets but also introduces a novel way of considering the importance of token dynamics, thereby enabling models to learn more pivotal representations.

\noindent\textbf{Limitations.} Although SBAM considers both the masking randomness and token salience, its primary focus on token dynamics may occasionally overlook the broader contextual nuances of less prominent tokens. Future work will explore the balance between focusing on highly salient tokens and developing methods that go beyond the masking randomness to consider tokens containing less significant yet subtle contexts, ensuring a more comprehensive image understanding.

\bibliographystyle{splncs04}
\bibliography{main}
\end{document}